\documentclass{article}
\usepackage{spconf,amsmath,epsfig}
\usepackage{booktabs} 
\usepackage{threeparttable}
\usepackage{epstopdf}
\usepackage{graphicx}
\usepackage{multicol}
\usepackage{multirow}
\usepackage[tight,footnotesize]{subfigure}
\usepackage{algorithm}
\usepackage{algpseudocode}
\usepackage{amsmath}
\usepackage{graphicx}
\usepackage{subfigure}   
\usepackage{cite}
\usepackage{amsfonts,amssymb}

\title{Attention Model Enhanced Network for Classification \\ of Breast Cancer Image }
\name{Xiao Kang$^{1}$, Xingbo Liu$^{1*}$, Xiushan Nie$^{2}$, Xiaoming Xi$^{2}$, Yilong Yin$^{1*}$ \thanks{Xingbo Liu and Yilong Yin are both the corresponding authors of this work.}}
\address{$^{1}$School of Software, Shandong University, Jinan, P.R. China\\
$^{2}$School of Computer Science and Technology, Shandong Jianzhu University , Jinan, P.R. China\\
\{sckx,sclxb\}@mail.sdu.edu.cn, niexsh@hotmail.com, fyzq10@126.com, ylyin@sdu.edu.cn      
}
\begin{document}

\maketitle

\begin{abstract}
Breast cancer classification remains a challenging task due to inter-class ambiguity and intra-class variability.  Existing deep learning-based methods try to confront this challenge by utilizing complex nonlinear projections. However, these methods typically extract global features from entire images, neglecting the fact that the subtle detail information can be crucial in extracting discriminative features. In this study, we propose a novel method named \underline{A}ttention \underline{M}odel \underline{E}nhanced \underline{N}etwork (AMEN), which is formulated in a multi-branch fashion with pixel-wised attention model and classification submodular. Specifically, the feature learning part in AMEN can generate pixel-wised attention map, while the classification submodular are utilized to classify the samples. To focus more on subtle detail information, the sample image is enhanced by the pixel-wised attention map generated from former branch.  Furthermore, boosting strategy are adopted to fuse classification results from different branches for better performance. Experiments conducted on three benchmark datasets demonstrate the superiority of the proposed method  under various scenarios.
\end{abstract}


\section{Introduction}

Breast cancer is the fifth most fatal disease and the second leading cause of death for women around the world \cite{antonio2019computer}.
The cause of this disease remains unknown, making prevention tricky. Thus, early detection and diagnosis, which can increase the success of treatment, reduce cost and even save lives, are strongly advocated. During the early diagnosis, radiologists prefer safer and more convenient imaging procedures, such as digital mammography and ultrasound-imaging, to detect tumors. However, medical image recognition is a  specialized and burdensome task depended on the experience of the pathologists. With the development of machine learning and  medical imaging technology, computer-assisted diagnosis, which can provide significant assistance for clinicians, attracts extensive attention in breast cancer classification field.

Differing from general coarse-grained natural image recognition, breast cancer classification suffers form huge inter-class ambiguity and intra-class variability \cite{szegedy2015GoogLeNet,huang2017DenseNet,He2016ResNet,fu2017look,liu2018fast,liu2018modality,liu2019SSLH,liu2019SDHMLR,liu2020model}. In addition, the training samples is rather insufficient due to the increasing cost of data collection and data privacy issues, hindering further improvement of computer-assisted diagnosis.
Therefore, it is momentous to acquire discriminative features in such a depressing condition. Previously, methods based on shallow features\cite{moon2011breast, duc2019classification} and handcraft features \cite{huang2020combining, irshad2013methods} have been extensively investigated. By designing manual features, Huang et al.\cite{huang2020combining} demonstrate that breast cancer recognition highly relative to exquisite information such as shape, orientation, margin integrality and so on. Although this method can gain satisfactory performance, it is considerably labor-intensive and extremely referred to the experience of the domain experts.

Recently, deep neural network has shown convincing efficacy in feature extraction, thus increasing works based on deep learning have been proposed in the field of medical image processing. Existing methods usually extract features from the entire image \cite{khan2019BCDeep, wu2019BCDeepNan, kaushiki2019patch}, which may lose subtle detail information of the lesion. To tackle this problem, Kaushiki et al. \cite{kaushiki2019patch} propose a patch-based classifier, which divides images into small patches and learns the features of each patch utilizing deep neural networks separately. Nevertheless, many patches seem not to have sufficient representational ability owing to intra-class variability of medical images. In other words,  similar patches tend to appear in different categories, making it arbitrary to assign the given semantic label to all patches \cite{2017Joint}. 

To address the aforementioned issues, we propose a multi-branch method for breast cancer classification named \underline{A}ttention \underline{M}odel \underline{E}nhanced \underline{N}etwork (AMEN) in this study. Two mutual tasks, {\emph{i.e.}} pixel-wised attention and classification, are realized in each branch. To capture more detail information,  pixel-wised attention map is utilized to reinforce the original images. Benefited by the iterative framework and  boosting strategy, the proposed method can achieve state-of-the-art performance on three benchmark datasets. The main contributions of this work are three-fold:

\begin{itemize}
\item \emph{Pixel-wised attention model}. Pixel-wised attention map generated from the feature learning part is supposed to indicate the importance of image pixels and capture the subtle detail information. 

\item \emph{Iterative framework and  boosting strategy}. The iterative process and  boosting strategy are conducive to achieve more stable and satisfactory precision.

\item  Experiments conducted on three benchmark datasets verify  the  superiority of the proposed method under various scenarios.
\end{itemize}

\section{The Proposed Method}
\begin{figure}[htp]
\centering\includegraphics[width=0.49\textwidth]{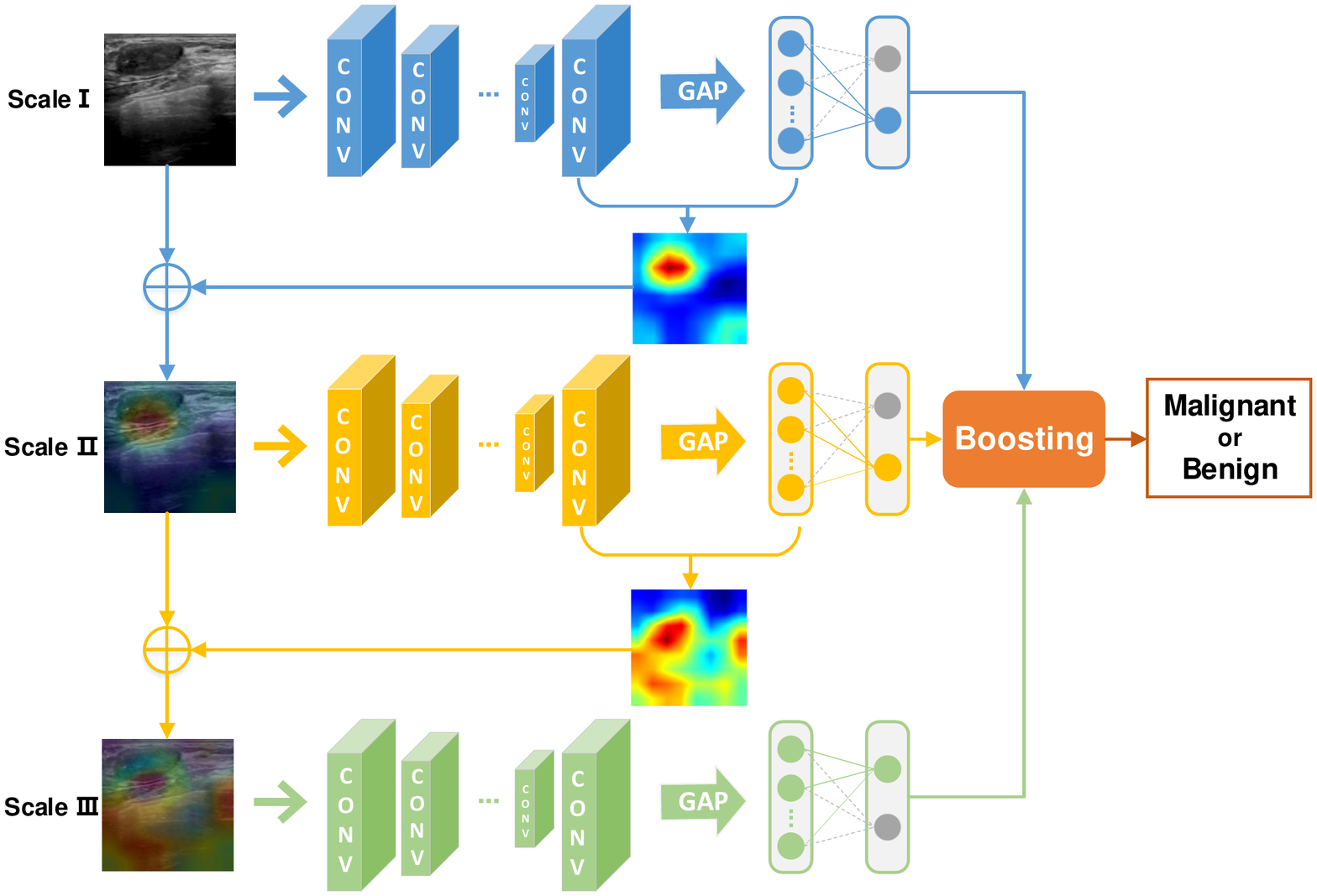}
\caption{The framework of the proposed AMEN when the number of branch is three. Global Average Pooling (GAP) is utilized to generate pixel-wised attention map  while $\bigoplus $ can enhance the original image with weighted pixel-wised attention map. Individual prediction in each branch is fused by the boosting strategy to achieve better performance.}\label{framework}
\end{figure}

This section elaborates the proposed method form several aspects, {\emph{i.e.}} feature
learning, classification submodular, pixel-wised attention model, iterative process and  prediction strategy. Fig. \ref{framework} shows the flowchart of the proposed AMEN. 

\subsection{Feature Extraction and Classification Network}

Assume that there is a training set $\bf{X}$ consisting of $N$ images, {\emph{i.e.}}, ${\bf{X}} = \{ {{\bf{x}}_n} \} _{n = 1}^N$, with ${{\bf{x}}_n}$ being the $n_{th}$ image. Additionally, the semantic label ${{\bf{y}}_n}\in $ $\{0,1\}^{M}$ is available, where $M$ is the number of categories. Specifically, ${y_{nm}} = 1$ if ${{\bf{x}}_n}$ belongs to class $m$ and 0 otherwise. 

To get the feature map ${\bf{F}}$ of the given image $\bf{X}$, deep neural network is utilized to preform feature extraction, and this process can be formulated as
\begin{equation}
  {\bf{F}} = D({\bf{X}}; \Theta_{d}),
\label{feature}
\end{equation}
where ${\bf{F}} \in \mathbb{R}^{W \times H \times C}$, $W$, $H$ and $C$ denote width, height and channel of the feature map, respectively. $\Theta_{d}$ is the parameter of  deep neural network while  $D(\cdot)$ is the combination of convolution, pooling and activation layers. The feature map ${\bf{F}}$ can be utilized in the following  two imperative parts of this study, {\emph{i.e.}} pixel-wised attention model and classification submodular.

To obtain reliable classification performance, we minimize the cross-entropy loss between the label vector $y_{n}$ and predicted probability $\sigma_{n}$ generated from classification network, which can be formalized as 
\begin{equation}
\begin{split}
&\min \limits_{{\bf{\Theta}}_{d}, {\bf{\Theta}}_{f}} -\frac{1}{N}\sum \limits _{i=1}^{N}\sum \limits _{j=1}^{M} y_{nm} \log (\sigma_{nm}), \\
&\textup{s.t.} \quad \sigma =softmax[fc({\bf{F}}; {\Theta}_{f} )].\\
\end{split}
\label{loss1}
\end{equation}
${\Theta}_{f}$ is the parameters for classification submodular, while $fc(\cdot)$ represents fully-connected layers. Moreover, the softmax function is adopted to  transform the outputs to classification probabilities. 

\begin{table*}[htp]  
  \centering  
  \fontsize{9}{11}\selectfont  
  \begin{threeparttable}  
  \caption{Overall comparison of four evaluation matrices on the three datasets. The OA, Sen, PPV and F1 indicate the Overall Accuracy, Sensitivity, Positive Predictive Value and F1-Score, respectively. The best results are shown in bold. }  
  \label{comparison}  
    \begin{tabular}{c|cccc|cccc|cccc}  
    \toprule  
    \multirow {2}{*}{Method} &\multicolumn{4}{c}{BCU} &\multicolumn{4}{|c}{BreaKHis 100X} &\multicolumn{4}{|c}{BreaKHis 200X}\cr
    \cmidrule(lr){2-5} \cmidrule(lr){6-9} \cmidrule(lr){10-13} 
    &OA&Sen&PPV&F1 &OA&Sen&PPV&F1 &OA&Sen&PPV&F1\cr  
    \midrule  
VGG16&0.7667&\bf0.9333&0.7000&0.8000&0.6728&0.5517&0.6337&0.5899&0.6690&0.6406&0.6357&0.6381\cr
VGG19&0.7667&0.8667&0.7222&0.7879&0.6728&0.6724&0.6047&0.6367&0.6963&0.6406&0.6949&0.6667\cr
GoogLeNet&0.7667&0.8000&0.7500&0.7742&0.8823&0.8321&\bf0.9268&0.8769&0.9087&0.8915&0.9349&0.9127\cr
Inception-v3&0.7333&0.7333&0.7333&0.7333&0.6765&0.5508&0.6500&0.5963&0.7469&0.6774&0.8000&0.7336\cr
DenseNet121&0.8000&0.8000&0.8000&0.8000&0.8419&0.8456&0.8342&0.8425&0.8548&0.8682&0.8615&0.8649\cr
ResNet50&0.7333&0.7333&0.7333&0.7333&0.8419&0.7794&0.8908&0.8314&0.8921&0.8702&0.9268&0.8976\cr
ResNet101&0.7667&0.7333&0.7857&0.7586&0.8272&0.7481&0.8860&0.8112&0.8963&0.8906&0.9120&0.9012\cr
\hline
${\textup{AMEN}}_1$ &0.8333&0.8667&0.8125&0.8387&\bf0.8897&0.8686&0.9084&\bf0.8881&\bf0.9294&0.9153&\bf0.9520&\bf0.9333\cr
${\textup{AMEN}}_2$ &\bf 0.8667&0.8667&\bf0.8667&\bf0.8667&0.8860&\bf0.8815&0.8881&0.8848&0.9170&\bf0.9448&0.9022&0.9230\cr
    \bottomrule  
    \end{tabular}  
   \end{threeparttable}  
\end{table*} 

\subsection{Pixel-wised Attention Model}
Attention model, inspired by region detection \cite{lin2013network}, can calculate pixel-wised attention map where high values indicate the essential pixels as far as classification is concerned. It is worth noting that this model are same in each branch.

Taking Scale I in Fig. \ref{framework} as an example, pixel-wised attention map ${\bf{A}} \in \mathbb{R} ^ {W \times H}$ is calculated by
\begin{equation}
 \label{GAP}
  {\bf{A}}_{w,h}= \sum \limits_{c=1}^{C}g({\bf{F}}_{w, h, c})\cdot {\bf{F}}_{w, h, c},
\end{equation}
where ${\bf{F}}_{w, h, c}$ means the element with coordinate ($w$, $h$, $c$) in feature map ${\bf{F}}$. $g(\cdot)$ denotes global average pooling \cite{lin2013network}, which can  transform feature map ${\bf{F}} \in \mathbb{R}^{W \times H \times C}$ to attention map ${\bf{A}}\in \mathbb{R}^{W \times H}$.  

Attention map ${\bf{A}}$ formulated in Eq. \ref{GAP}, has the same size with original image, with each value in ${\bf{A}}$ indicating the importance of the pixel corresponding to original image. Compared with \cite{kaushiki2019patch}, the attention map in this study is pixel-wised, making the subtle detail information more flexible and accurate.
Intuitively, regions with high values can be identified as discriminative parts that are vital for classification task. 

\subsection{Overall Structure}
In this section, we will introduce the overall structure of the proposed AMEN by taking a three-branch framework in Fig. \ref{framework} for example. In detail, the classification process consists of the following three steps:

Step 1: we initialize the network in first branch ({\emph{i.e.}} Scale I) by the pretrained backbone from ImageNet \cite{olga2015imagenet}. Then, we train the feature extraction network and classification submodule simultaneously with the provided dataset ${\bf{X}}$ by back propagation. Next, pixel-wised attention map ${\bf{A}}_{1}$ is generated from feature learning part. And prediction results are saved for boosting strategy.

Step 2: we initialize network of next branch by the parameters in previous scale. To capture more detail information, we enhance the original image with the learned pixel-wised attention map. We adopt a plain but effective weighted superposition strategy in this study, and this process can be formalized as
\begin{equation}
    {\bf{X}}_{s}={\bf{X}}_{s-1}+\lambda_{s-1} {\bf{A}}_{s-1},
    \label{input_image}
\end{equation}
where ${\bf{X}}_{s}$ denotes the input image of $s_{th}$ branch,  and  $\lambda_{s-1}$ is a hyperparameter which balances original image and attention map. Then the feature extraction network and classification submodule are trained with the enhanced dataset ${\bf{X}}_{s}$ by back propagation. By this iterative strategy, we can stack more branches to get better performance, which will be shown in the experimental results.
In conclusion, the loss function in each branch can be defined as 
\begin{equation}  
\label{final_loss}
\begin{split}
&\min \limits_{{\bf{\Theta}}_{s,d}, {\bf{\Theta}}_{s,f}} -\frac{1}{N}\sum \limits _{n=1}^{N}\sum \limits _{m=1}^{M} y_{nm} \log (\sigma_{nm}), \\
&\textup{s.t.} \quad \sigma=softmax\{fc[D({\bf{X}}_{s};\Theta_{s,d} );{\Theta_{s,f}}]\},\\
\end{split}    
\end{equation}
where $s=1,2,\cdots, S$ , $\lambda_{1}=0$, ${\bf{A}}_{s-1}={\bf{0}}$ and ${\bf{X}}_{0}={\bf{X}}$.

Step 3: Enhanced by the learned pixel-wised attention map, the subsequent branches are supposed to achieve more distinguishing classification ability. Even so, to get more stable and practical classification performance, a simple but effective boosting strategy, {\emph{i.e.}} majority voting, is adopted to fuse the prediction results in each branch.

By utilizing the above three steps, the proposed method can be considered as a multi-branch boosting framework. It is noted that each independent branch can be seamlessly integrated by pixel-wised attention model and boosting strategy.

\begin{table*}[htp]  
  \centering  
  \fontsize{9}{11}\selectfont  
  \begin{threeparttable}  
  \caption{Ablation study of four evaluation matrices on the three datasets.   }  
  \label{Ablation}  
    \begin{tabular}{c|cccc|cccc|cccc}  
    \toprule  
    \multirow {2}{*}{Method} &\multicolumn{4}{c}{BCU} &\multicolumn{4}{|c}{BreaKHis 100X} &\multicolumn{4}{|c}{BreaKHis 200X}\cr
    \cmidrule(lr){2-5} \cmidrule(lr){6-9} \cmidrule(lr){10-13} 
    &OA&Sen&PPV&F1 &OA&Sen&PPV&F1 &OA&Sen&PPV&F1\cr  
    \midrule  
Average &0.7444&0.7556&0.7389&0.7470&0.8767&0.8415&0.9082&0.8734&0.9030&0.8977&0.9148&0.9071\cr
Boosting&0.7667&0.8000&0.7500&0.7742&0.8823&0.8321&0.9268&0.8769&0.9129&0.9206&0.9134&0.9170\cr
Scale I&0.7667&0.8000&0.7500&0.7742&0.8713&0.8248&0.9262&0.8726&0.9087&0.8915&0.9349&0.9127\cr
Scale II&0.8000&0.8667&0.7647&0.8125&0.8787&0.8540&0.9000&0.8764&0.9129&0.9008&0.9365&0.9183\cr
Scale III&0.8000&0.8667&0.7647&0.8125&0.8823&0.8321&0.9268&0.8769&0.9212&0.9077&0.9440&0.9255\cr
\hline
${\textup{AMEN}}_1$ &0.8333&0.8667&0.8125&0.8387&0.8897&0.8686&0.9084&0.8881&0.9294&0.9153&0.9520&0.9333\cr
\midrule[0.8pt]
Average&0.7667&0.7778&0.7611&0.7692&0.8368&0.8391&0.8280&0.8343&0.8562&0.8574&0.8668&0.8620\cr
Boosting&0.8000&0.8667&0.7647&0.8125&0.8548&0.8793&0.8293&0.8536&0.8631&0.8571&0.8780&0.8675\cr
Scale I&0.8000&0.8000&0.8000&0.8000&0.8419&0.8456&0.8342&0.8425&0.8548&0.8682&0.8615&0.8649\cr
Scale II&0.8333&0.8000&0.8571&0.8276&0.8529&0.8129&0.8898&0.8496&0.8755&0.8837&0.8837&0.8837\cr
Scale III&0.8333&0.8000&0.8571&0.8276&0.8713&0.8322&0.9048&0.8669&0.9087&0.9077&0.9219&0.9147\cr
\hline
${\textup{AMEN}}_2$ &0.8667&0.8667&0.8667&0.8667&0.8860&0.8815&0.8881&0.8848&0.9170&0.9448&0.9022&0.9230\cr
    \bottomrule  
    \end{tabular}  
   \end{threeparttable}  
\end{table*} 

\begin{figure}[htb]
\centering
\subfigure[BCU]{
\includegraphics[width=0.15\textwidth]{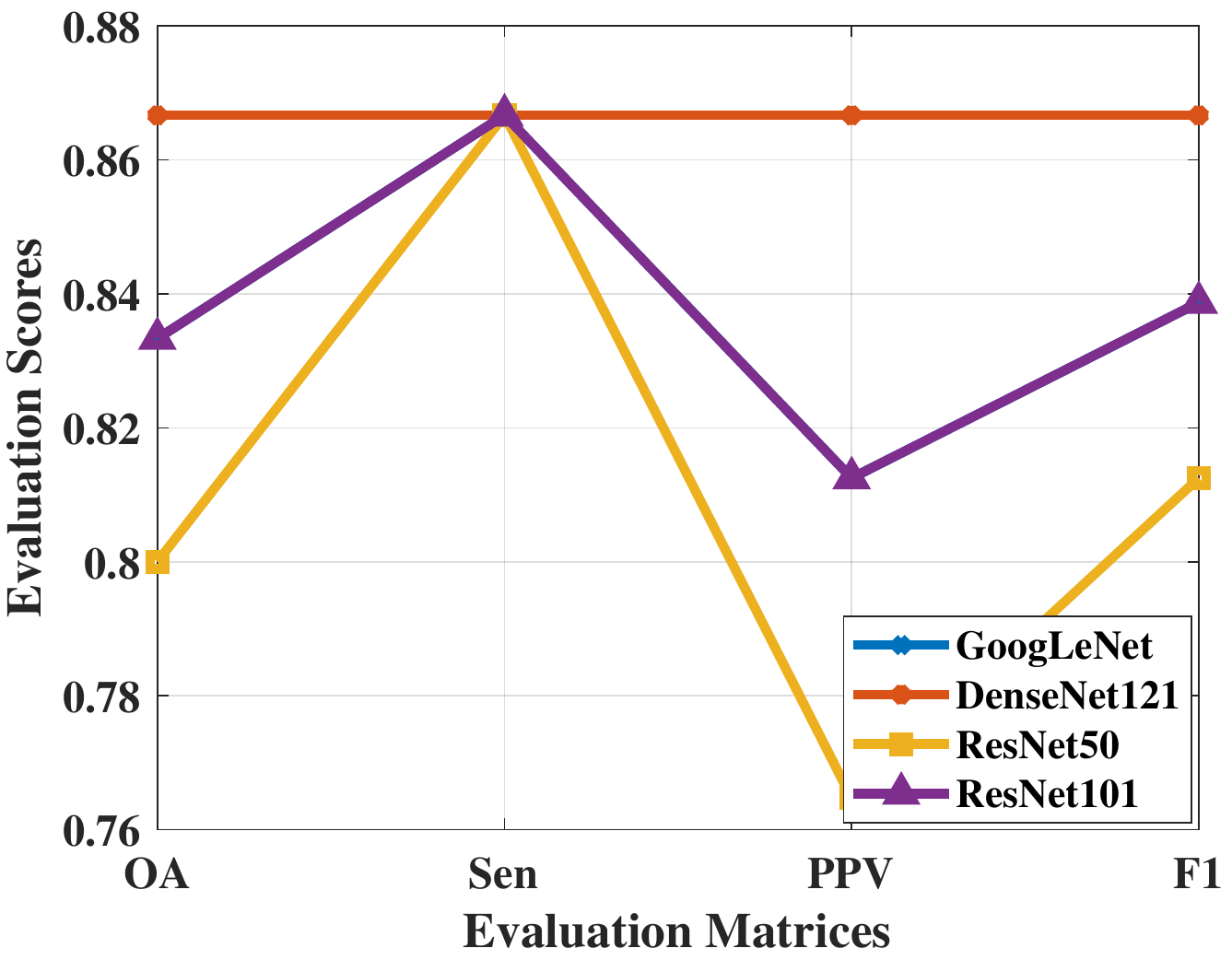}}
\subfigure[BreaKHis 100X]{
\includegraphics[width=0.15\textwidth]{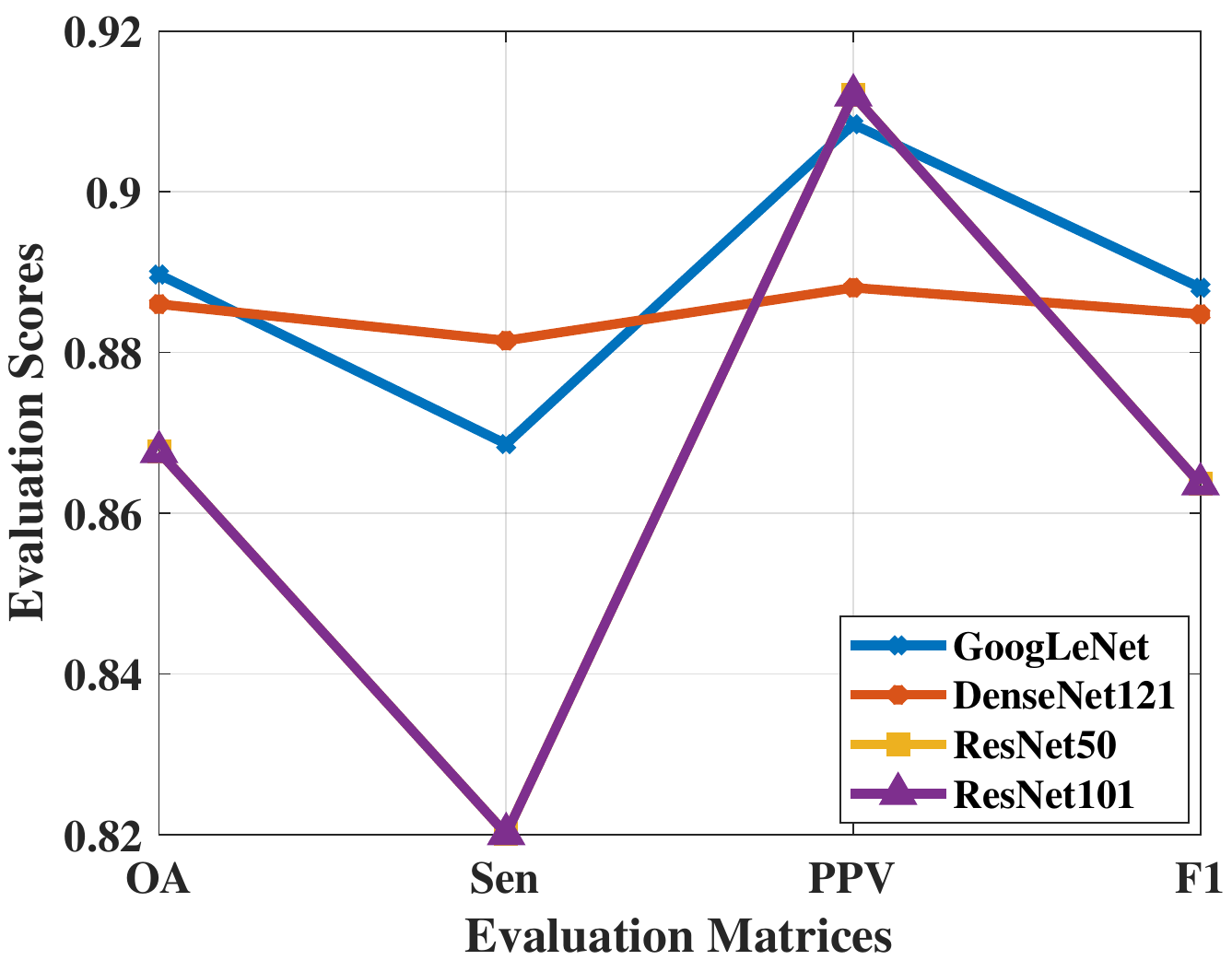}}
\subfigure[BreaKHis 200X]{
\includegraphics[width=0.15\textwidth]{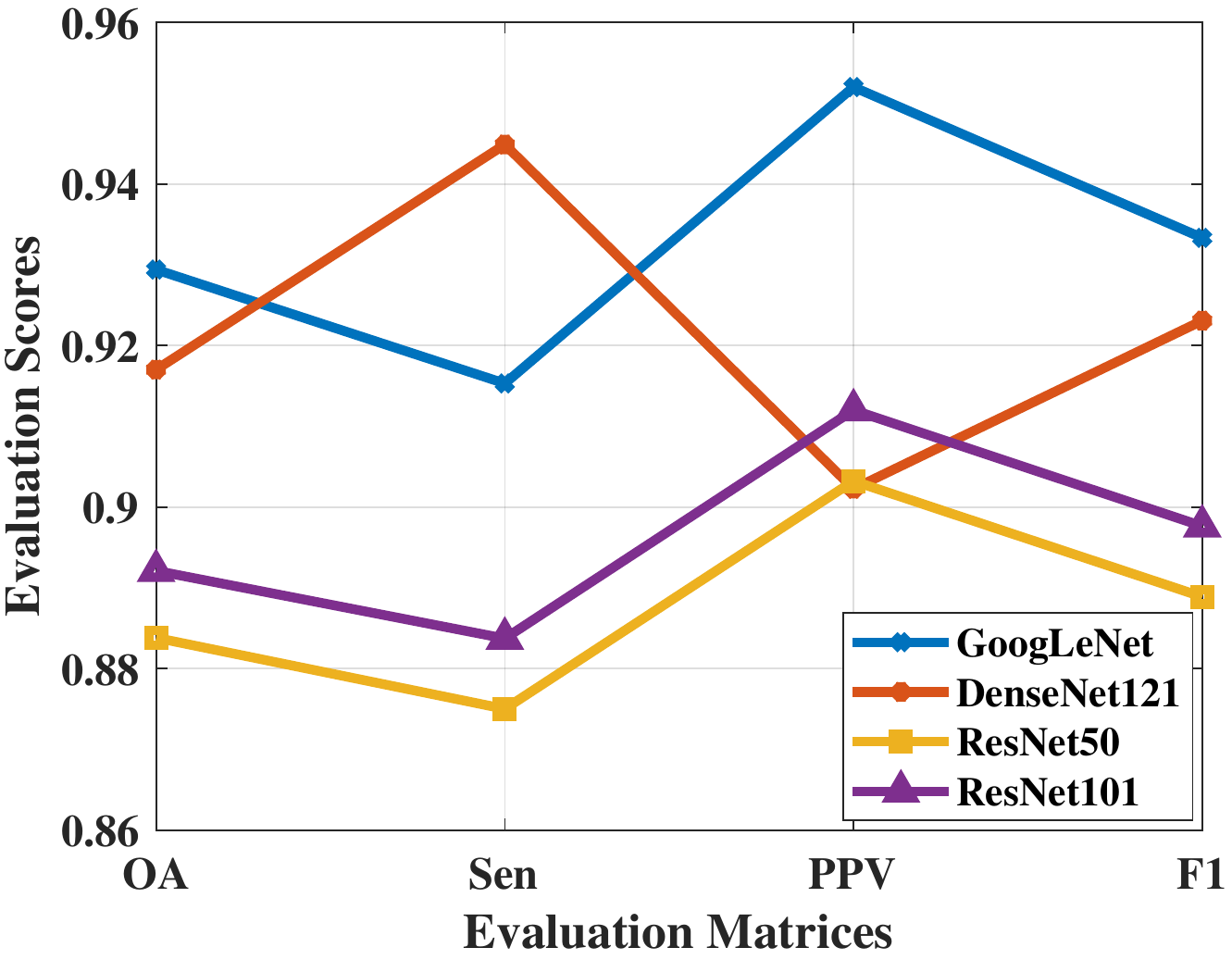}}
\caption{Performance in terms of four evaluation matrices when utilizing different backbones based on three datasets.  }\label{Boosting}
\end{figure}

\begin{figure}[htb]
\centering
\subfigure[BCU]{
\includegraphics[width=0.15\textwidth]{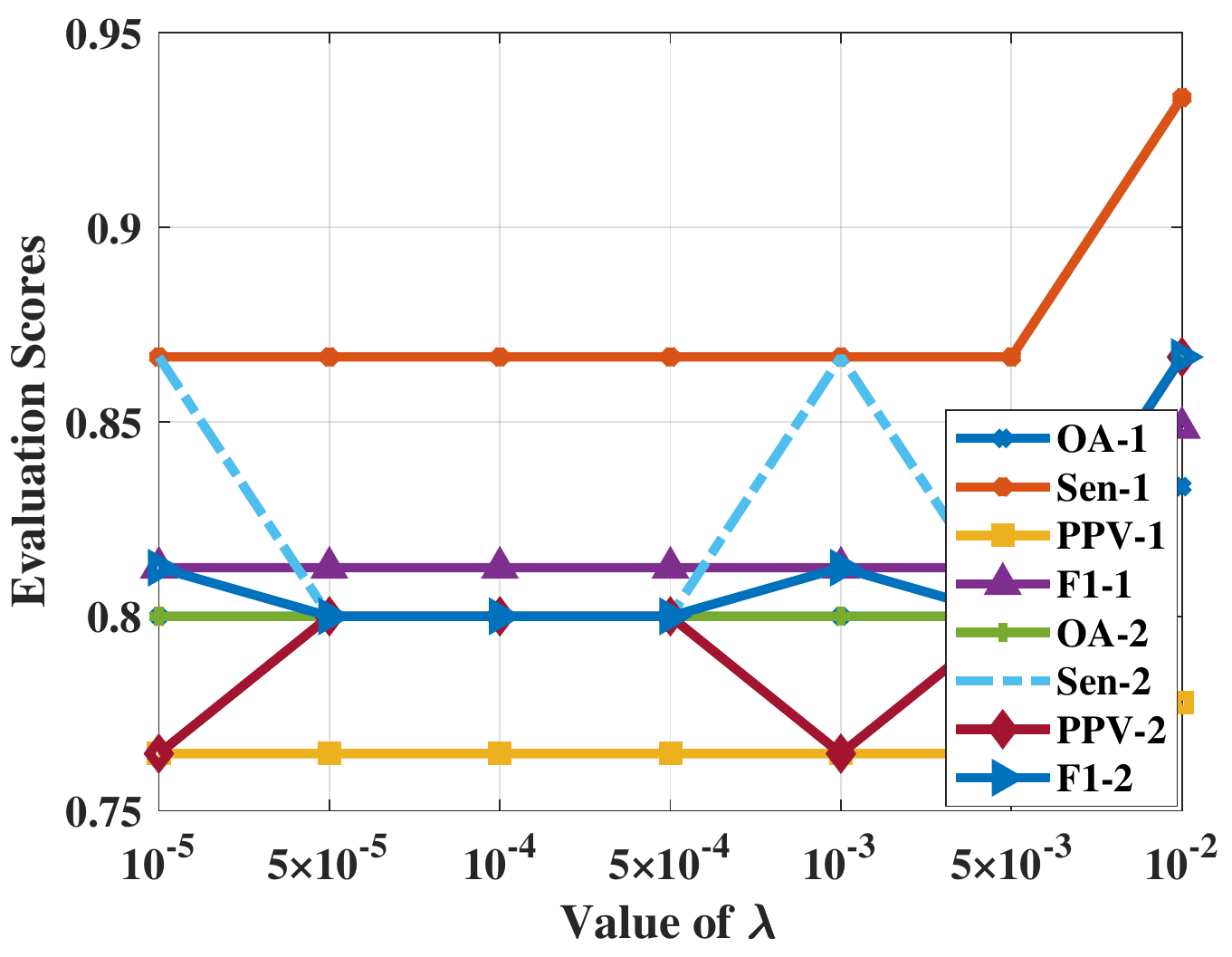}}
\subfigure[BreaKHis 100X]{
\includegraphics[width=0.15\textwidth]{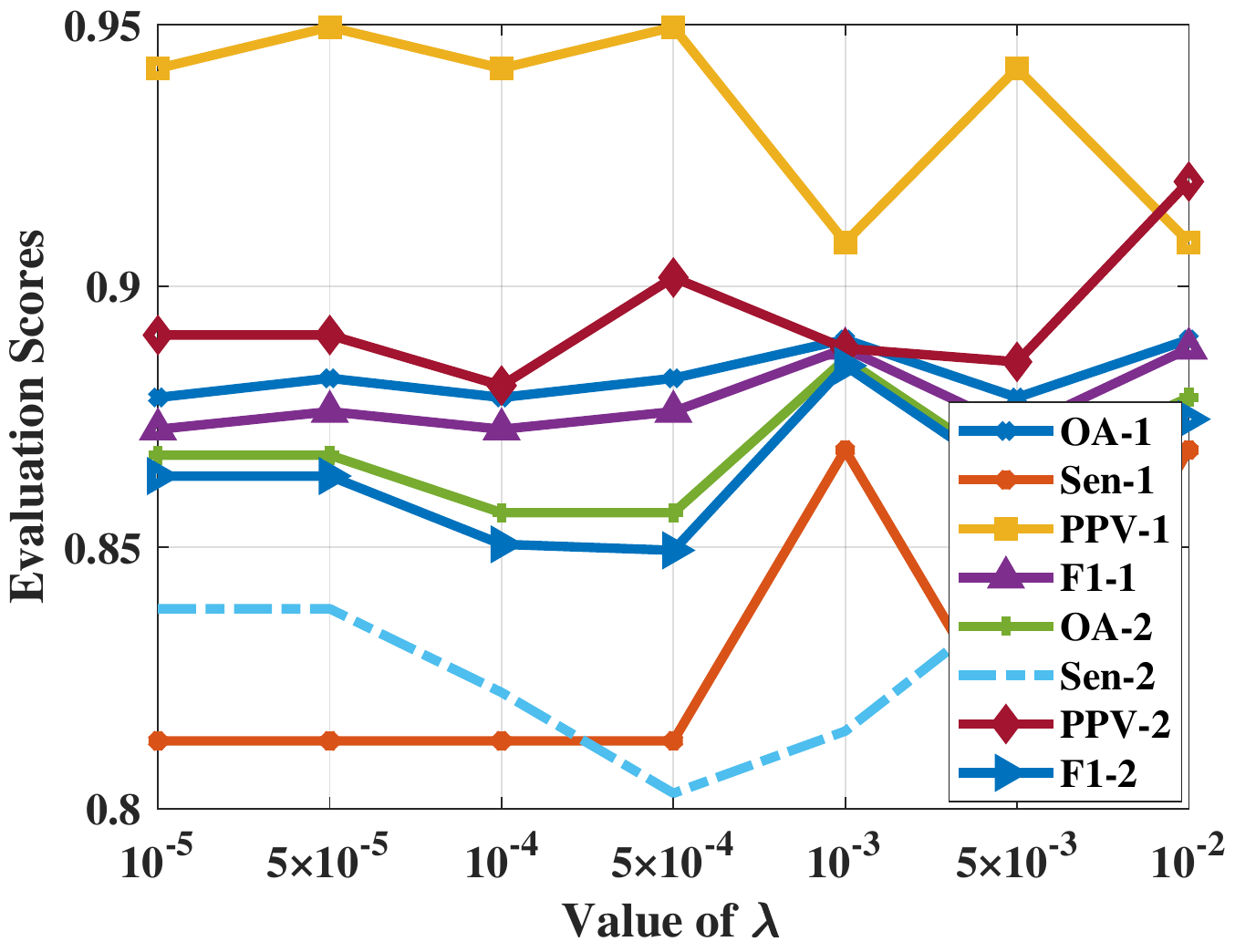}}
\subfigure[BreaKHis 200X]{
\includegraphics[width=0.15\textwidth]{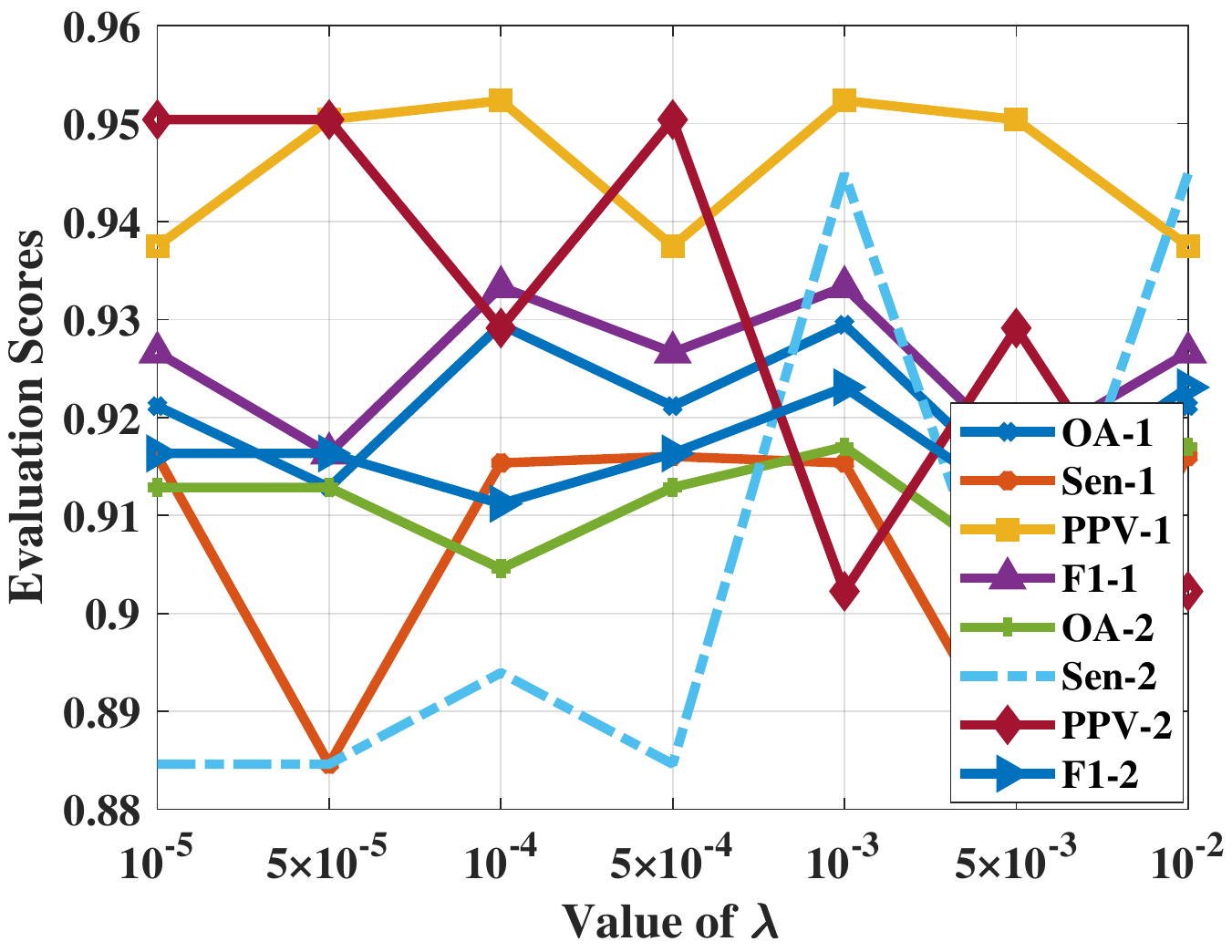}}
\caption{Performance in terms of four evaluation matrices when $\lambda$ varies in a range based on three  benchmark datasets. The OA-1 and OA-2 represent the Overall Accuracy of the proposed ${\textup{AMEN}}_1$ and ${\textup{AMEN}}_2$, respectively. }\label{PPS}
\end{figure}

\section{EXPERIMENT}
\subsection{Experimental Settings}
To confirm the superiority of the proposed method, we conduct experiments on three benchmark datasets, Breast Cancer Ultrasound (BCU)\cite{wu2019BCUDataSet}, BreaKHis 100X and BreaKHis 200X\cite{fabio2016BreaKHis}. The BCU dataset is a ultrasound image dataset while BreaKHis 100X and BreaKHis 200X are histology image datasets. In this study, we compare the proposed AMEN with Visual Geometry Group Network (VGG16 and VGG19) \cite{simonyan2014VGG}, Inception-v3 \cite{christian2016inception}, Deep Residual Network (ResNet50 and ResNet101) \cite{He2016ResNet}, GoogLeNet \cite{szegedy2015GoogLeNet}, and Dense Convolutional Network (DenseNet121) \cite{huang2017DenseNet}. \emph{Overall accuracy}, \emph{Sensitivity}, \emph{Positive predictive value},  and \emph{F1-Score} \cite{heng2010Criterion} are adopted to evaluate the performance of these methods on the three datasets. 

The number of scales is set to 3 empirically in this study and more can be stacked in a similar way. For the hyperparameter ${\lambda}_{s}$ in the Eq. \ref{input_image}, we set it as $10^{-3}$ in each scale for  simplicity. 
Empirically, values of epoch, learning rate, momentum and weight decay are set to 100, $10^{-4}$, 0.99 and $10^{-2}$, respectively. The images of three benchmark datasets are resized to 256 $\times$ 256.
Pytorch are used as the code-base, and all the models are trained on 4 TITAN X GPUs. More details to our code is presented in https://github.com/......
   
\subsection{Experimental Results and Analysis}
According to the performance on different backbone, we take the classification network with better accuracy, {\emph{i.e.}}, GoogLeNet  and DenseNet121, as backbones of the proposed AMEN. As shown in Table \ref{comparison}, the proposed AMEN$_{1}$ (AMEN with GoogLeNet) and AMEN$_{2}$ (AMEN with DenseNet121) are superior to the compared methods in nearly all of the evaluation criterion with an improvement of around 3\%. 

Table \ref{Ablation} presents the results of ablation experiments. We run the backbones for three times and list the average performance (shown as Average). In addition, we fuse the three results using the same boosting strategy (shown as Boosting). Furthermore, we exhibit the performance of the proposed AMEN with three branches (shown as Scale I, Scale II and Scale III).  We have a primary observation that methods based on boosting strategy can achieve better performance.
Furthermore, the branches which take pixel-wised attention map into consideration are more discriminative for classification.  Additionally, the performance seems to be improved as the number of scales increasing.  Last but not least, the proposed method with both pixel-wised attention and boosting strategy outperforms on the three benchmark datasets, verifying the efficiency and superiority.

Fig. \ref{Boosting} exhibits the performance  in terms of four evaluation matrices when utilizing four backbones, ResNet50, ResNet101, GoogLeNet and DenseNet121. The proposed AMEN with GoogLeNet and DenseNet121 can gain better performances compared with their counterparts. 
In addition, Fig. \ref{PPS} shows the performance in terms of four evaluation matrices when ${\lambda}_{s}$ ranging form $10^{-5}$ to $10^{-2}$. The proposed method exhibit acceptable stability with different granularity of ${\lambda}_{s}$, saving lots of time for parameter tuning.

\section{CONCLUSION}
In this study, we propose a deep learning-based method for breast cancer classification called Attention Model Enhanced Network, which integrates pixel-wised attention and multi-branch boosting in an iterative fashion. The pixel-wised attention can remedy the intractable lost of subtle detail information by a simple but effective weighted superposition strategy. Furthermore, the multi-branch boosting is conducive to achieve more stable and compelling performance. Experiments conducted on three benchmark datasets validate the superiority of the proposed method under various scenarios.

\section{ACKNOWLEDGEMENTS}
This work was supported in part by the National Natural Science Foundation of China (61876098, 61671274, 61573219), National Key R$\&$D Program of China (2018YFC0830100, 2018YFC0830102) and special funds for distinguished professors of Shandong Jianzhu University.
\bibliographystyle{IEEEbib}
\bibliography{refs}
\end{document}